\documentclass[sigconf]{acmart}  

\usepackage{tikz}
\usepackage{pgfplots}
\usepackage{xcolor}
\usepackage{filecontents}

\usepackage{amsmath}
\usepackage{graphicx}
\usepackage{url}
\usepackage{hyperref}
\usepackage{subfig}
\usepackage{amsfonts}
\usepackage{color, soul}
\usepackage{mathrsfs}
\usepackage{cleveref}
\usepackage{multirow}
\usepackage{float}
\usepackage{wrapfig}
\usepackage{amsthm}
\usepackage{bm}
\usepackage{bbm}
\usepackage{algorithm}
\usepackage{algorithmic}
\usepackage{colortbl}
\usepackage{enumitem}
\usepackage[T1]{fontenc}
\usepackage{aecompl}
\usepackage{dsfont}

\newtheorem{myDef}{Definition}
\newtheorem{myDef1}{Problem}

\AtBeginDocument{%
  \providecommand\BibTeX{{%
    \normalfont B\kern-0.5em{\scshape i\kern-0.25em b}\kern-0.8em\TeX}}}

\copyrightyear{2022} 
\acmYear{2022} 
\setcopyright{acmcopyright}\acmConference[KDD '22]{Proceedings of the 28th ACM SIGKDD Conference on Knowledge Discovery and Data Mining}{August 14--18, 2022}{Washington, DC, USA}
\acmBooktitle{Proceedings of the 28th ACM SIGKDD Conference on Knowledge Discovery and Data Mining (KDD '22), August 14--18, 2022, Washington, DC, USA}
\acmPrice{15.00}
\acmDOI{10.1145/3534678.3539319}
\acmISBN{978-1-4503-9385-0/22/08}

\begin{document}


\title{On Structural Explanation of Bias in Graph Neural Networks}


\vspace{-4ex}
\author{Yushun Dong}
\email{yd6eb@virginia.edu}
\affiliation{%
  \institution{University of Virginia}
  \country{}
}

\author{Song Wang}
\email{sw3wv@virginia.edu}
\affiliation{%
  \institution{University of Virginia}
  \country{}
}

\author{Yu Wang}
\email{yu.wang.1@vanderbilt.edu}
\affiliation{%
  \institution{Vanderbilt University}
  \country{}
}

\author{Tyler Derr}
\email{tyler.derr@vanderbilt.edu}
\affiliation{%
  \institution{Vanderbilt University}
  \country{}
}

\author{Jundong Li}
\email{jundong@virginia.edu}
\affiliation{%
  \institution{University of Virginia}
  \country{}
}

\begin{abstract}

Graph Neural Networks (GNNs) have shown satisfying performance in various graph analytical problems. Hence, they have become the \emph{de facto} solution in a variety of decision-making scenarios. However, GNNs could yield biased results against certain demographic subgroups. Some recent works have empirically shown that the biased structure of the input network is a significant source of bias for GNNs.
Nevertheless, no studies have systematically scrutinized which part of the input network structure leads to biased predictions for any given node. 
The low transparency on how the structure of the input network influences the bias in GNN outcome largely limits the safe adoption of GNNs in various decision-critical scenarios. 
In this paper, we study a novel research problem of structural explanation of bias in GNNs. Specifically, we propose a novel post-hoc explanation framework to identify two edge sets that can maximally account for the exhibited bias and maximally contribute to the fairness level of the GNN prediction for any given node, respectively. Such explanations not only provide a comprehensive understanding of bias/fairness of GNN predictions but also have practical significance in building an effective yet fair GNN model. Extensive experiments on real-world datasets validate the effectiveness of the proposed framework towards delivering effective structural explanations for the bias of GNNs. 
Open-source code can be found at \href{https://github.com/yushundong/REFEREE}{https://github.com/yushundong/REFEREE}.

\vspace{-0.75ex}
\end{abstract}

\keywords{\vspace{-2.5ex}\\
Graph neural networks, algorithmic fairness, model explanability\vspace{-0ex}}

\ccsdesc[500]{Information systems~Data mining}
\ccsdesc[500]{Computing methodologies~Machine learning\vspace{-1ex}}

\maketitle

\section{Introduction}
\label{intro}


Graph Neural Networks (GNNs) have shown satisfying performance in various real-world applications, e.g., online recommendation~\cite{wu2019session}, chemical reaction prediction~\cite{do2019graph}, and complex physics simulation~\cite{sanchez2020learning}, to name a few.
The success of GNNs is generally attributed to their message-passing mechanism~\cite{wu2020comprehensive,zhou2020graph,wang2021tree}. Such a mechanism enables GNNs to capture the correlation between any node and its neighbors in a localized subgraph (i.e., the computation graph of the node~\cite{ying2019gnnexplainer}), which helps to extract information from both node attributes and network structure for node embedding learning~\cite{hamilton2017inductive}.
Despite the remarkable success, most of the existing GNNs do not have fairness consideration~\cite{dong2021individual,dong2021edits,dai2021say,kang2022rawlsgcn,dong2022fairness,DBLP:conf/wsdm/MaGWYZL22}. 
Consequently, GNN predictions could exhibit discrimination (i.e., bias) towards specific demographic subgroups that are described by sensitive features, e.g., age, gender, and race.
Such discrimination has become one of the most critical societal concerns when GNNs are deployed in high-stake decision-making scenarios~\cite{kang2021fair}.

There is a rich body of literature on alleviating the bias of GNNs. Generally, these works aim to decouple the learned node embeddings from sensitive features~\cite{dai2021say,dong2021edits,DBLP:conf/iclr/LiWZHL21,spinelli2021biased,fairview}.
However, they cannot provide explanations on how bias arises in GNNs.
In fact, it is worth noting that in various high-stake decision-making scenarios, we not only need to alleviate bias in GNNs, but also need to understand how bias is introduced to the prediction of each individual data instance (e.g., a node in a graph).
Such instance-level understanding is critical for the safe deployment of GNNs in decision-critical applications~\cite{ying2019gnnexplainer}.
For example, GNNs have demonstrated superior performance in many financial applications, such as loan approval prediction for bank clients~\cite{wang2019semi,xu2021towards}. In this scenario, different clients form a graph based on their transaction interactions, and the records of clients form their features. Here, the goal is to predict whether a client will be approved for a loan, and such a problem can be formulated as a node classification task that can be solved by GNNs. However, GNNs could lead to undesired discrimination against clients from certain demographic subgroups (e.g., rejecting a loan request mainly because the applicant belongs to an underprivileged group). In this example, understanding how bias is introduced to the prediction of each individual client enables bank managers to scrutinize each specific loan decision and take proactive actions to improve the algorithm and reduce potential discrimination.

In fact, biased GNN predictions can be attributed to a variety of factors. Among them, biased network structure has shown to be a critical source~\cite{DBLP:conf/iclr/LiWZHL21,dong2021edits,spinelli2021biased}. Additionally, bias in the network structure could be amplified by the core operation of GNNs -- the \emph{message-passing} mechanism~\cite{dong2021edits}.
Therefore, understanding which part of the network structure leads to biased GNN predictions for each node is vitally important.
Towards this goal, we aim to provide an instance-level (i.e., node-level) structural explanation of bias in GNN predictions.
%
%
More specifically, for any node in an input network for GNNs, we aim to understand and explain how different edges in its computation graph contribute to the level of bias for its prediction\footnote{Here, we only consider the edges in its corresponding computation graph. This is because the computation graph of a node fully encodes all information that GNN models leverage to generate its prediction~\cite{ying2019gnnexplainer}.
}.
Nevertheless, it remains a daunting task. Essentially, we mainly face the following three challenges:
(1) \textbf{Fairness Notion Gap}: \emph{how to measure the level of bias for the GNN prediction at the instance level}? For each node, understanding how the edges in its computation graph make its prediction biased requires a principled bias metric at the instance level. However, most of the existing bias metrics are defined over the whole population or the sub-population~\cite{dwork2012fairness,DBLP:conf/nips/HardtPNS16}, thus they cannot be directly grafted to our studied problem. In this regard, it is crucial to design a bias metric that can quantify the level of bias for the GNN prediction at the instance level.
(2) \textbf{Usability Gap}: \emph{is a single bias explainer sufficient}? 
%
It should be noted that our ultimate goal goes beyond explaining bias as we also aim to achieve fairer GNNs, which provide better model usability and enable ethical decision-making. Consequently, it is also critical to explain which edges in a node's computational graph contribute more to the fairness level of its prediction. However, edges that introduce the least bias cannot be simply regarded as the edges that maximally contribute to the fairness level of the prediction. This is because edges that introduce the least bias could also be those prediction-irrelevant edges --- such edges could barely contribute any information to the GNN prediction. Therefore, only explaining how each edge in a computational graph contributes to the exhibited node-level bias is not sufficient. 
%
(3) \textbf{Faithfulness Gap}: \emph{how to obtain bias (fairness) explanations that are faithful to the GNN prediction}? 
To ensure the obtained explanations reflect the true reasoning results based on the given GNN model, most existing works on the instance-level GNN explanation obtain explanations that encode as much critical information as possible for a given GNN prediction~\cite{ying2019gnnexplainer,vu2020pgm,luo2020parameterized}. In this way, the obtained explanations are considered to be faithful to the given GNN model, as they generally reflect the critical information the GNN utilized to make the given prediction. 
%
Similarly, when explaining how the bias or the fairness level of the GNN prediction is achieved, we are also supposed to identify the critical information the GNN utilized to achieve such a level of bias or fairness for the given prediction.

As an attempt to tackle the challenges above, in this paper, we propose a principled framework named REFEREE (st\underline{R}uctural \underline{E}xplanation o\underline{F} bias\underline{E}s in g\underline{R}aph n\underline{E}ural n\underline{E}tworks) for post-hoc explanation of bias in GNNs. 
Specifically, towards the goal of obtaining instance-level structural explanations of bias,
we formulate a novel research problem of \textit{Structural Explanation of Node-Level Bias in GNNs}.
To tackle the first challenge, we propose a novel fairness notion together with the corresponding metric to measure the level of bias for a specific node in terms of GNN prediction.
To tackle the second challenge, we design two explainers in the proposed framework REFEREE, namely bias explainer and fairness explainer. In any given computation graph, they are able to identify edges that maximally account for the exhibited bias in the prediction and edges that maximally contribute to the fairness level of the prediction, respectively.
To tackle the third challenge, we design a constraint to enforce the faithfulness for the identified explanations, which can be incorporated into a unified objective function for the proposed framework. In this way, apart from the goal of explaining the exhibited bias and identifying edges that help with fairness, such a unified objective function also enforces the identified explanations to be faithful to the given GNN prediction.
To better differentiate these two types of edges, the two explainers are designed to work in a contrastive manner.
%
%
%
%
%
Finally, we evaluate the effectiveness of REFEREE on multiple real-world network datasets.
The main contributions of this paper are as follows.
(1) \textbf{\emph{Problem Formulation.}} We formulate and study a novel problem of structural explanation of biases in GNNs given any instance-level GNN prediction.
(2) \textbf{\emph{Metric and Algorithmic Design.}} We propose a novel metric to measure how biased the GNN outcome prediction of a node is. We then propose a novel explanation framework named REFEREE to provide explanations on both fairness and bias, and maintain faithfulness to the given prediction.
(3) \textbf{\emph{Experimental Evaluations.}} We perform experimental evaluations on various real-world networks. Extensive experiments demonstrate the effectiveness of REFEREE and its superiority over other alternatives.

\vspace{-1.0em}
\section{Problem Definition}

\vspace{-0.3em}


In this section, we first present the notations used in this paper and some preliminaries. We then formulate a novel problem of \textit{Structural Explanation of Bias in GNNs}.

\noindent \textbf{Notations.} We use bold uppercase letters (e.g., $\mathbf{A}$), bold lowercase letters (e.g., $\mathbf{x}$), and normal uppercase letters (e.g., $N$) to represent matrices, vectors, and scalars, respectively. 
Uppercase letters in math calligraphy font (e.g., $\mathcal{V}$) represent sets.
The $k$-th entry of a vector, e.g., $\mathbf{x}$, is represented as $\mathbf{x}[k]$. 
For any number, $|\cdot|$ is the absolute value operator; for any set, $|\cdot|$ outputs its cardinality.

\noindent \textbf{Preliminaries.} 
We denote an attributed network as $\mathcal{G} = \{\mathcal{V}, \mathcal{E}, \mathcal{X}\}$, where $\mathcal{V} = \{v_1, ..., v_N\}$ represents the set of $N$ nodes; $\mathcal{E} \subseteq \mathcal{V} \times \mathcal{V}$ is the set of all edges; $\mathcal{X} = \{\mathbf{x}_1, ..., \mathbf{x}_N\}$ is the set of node attribute vectors. 
A trained GNN model $f_{\bm{\Theta}}$ maps each node to the outcome space, where $\bm{\Theta}$ denotes the parameters of the GNN model. Without loss of generality, we consider node classification as the downstream task. The GNN outcome for $N$ nodes can be given as $\mathcal{\hat{Y}} = \{\hat{\mathbf{y}}_1, ..., \hat{\mathbf{y}}_i, ..., \hat{\mathbf{y}}_N\}$, where $ \hat{\mathbf{y}}_i \in \mathbb{R}^{C}$. Here $C$ is the number of classes for node classification, and each dimension in $\hat{\mathbf{y}}_i$ represents the probability of the node belonging to the corresponding class.
Based on $\mathcal{\hat{Y}}$, the predicted label set by GNN for these $N$ nodes is denoted by $\{\hat{Y}_1, ..., \hat{Y}_i, ..., \hat{Y}_N\}$. Here $\hat{Y}_i$ is determined by the highest predicted probability across all $C$ classes given by $\hat{\mathbf{y}}_i$.
For GNN explanation, we consider the most widely studied instance-level explanation problem in this paper, i.e., we aim to explain the given prediction of a node based on its computation graph~\cite{ying2019gnnexplainer,vu2020pgm,luo2020parameterized}.
%
At the instance level, the explanations can be provided from different perspectives.
Here we focus on the structural explanation, i.e., the explanation is given as an edge set $\mathcal{\tilde{E}}_i$ by any GNN explanation model $h_{\bm{\Phi}}$.
Specifically, given a specific node $v_i$, its computation graph $\mathcal{G}_i = \{\mathcal{V}_i, \mathcal{E}_i, \mathcal{X}_i\}$ (i.e., the $L$-hop subgraph centered on node $v_i$~\cite{ying2019gnnexplainer}, where $L$ is the total layer number of the studied GNN), and the corresponding outcome $\hat{\textbf{y}}_i$, the GNN structural explanation model $h_{\bm{\Phi}}$ with parameter $\bm{\Phi}$ identifies an explanation as an edge set $\mathcal{\tilde{E}}_i$ corresponding to the outcome $\hat{\textbf{y}}_i$.
%
$\mathcal{\tilde{E}}_i$ is identified through learning a weighted mask matrix $\mathbf{M} \in \mathbb{R}^{|\mathcal{V}_i| \times |\mathcal{V}_i|}$ that indicates the importance score of each edge in $\mathcal{E}_i$. 
Edges in $\mathcal{\tilde{E}}_i$ are selected from $\mathcal{E}_i$ based on such importance score.
%
%
We denote the computation graph with the identified edge set $\mathcal{\tilde{E}}_i$ as a new subgraph $\mathcal{\tilde{G}}_i = \{\mathcal{V}_i, \mathcal{\tilde{E}}_i, \mathcal{X}_i\}$.
Based on the new subgraph $\mathcal{\tilde{G}}_i$ with the identified edge set $\mathcal{\tilde{E}}_i$, the given GNN yields a different probabilistic outcome $\tilde{\textbf{y}}_i = f_{\bm{\Theta}} (\mathcal{\tilde{G}}_i)$ compared with the vanilla outcome $\hat{\textbf{y}}_i$.
Based on the above notations and preliminaries, we formulate the problem of \textit{Structural Explanation of Bias in GNNs} as follows.

\begin{myDef1}
\label{p1}
\textbf{Structural Explanation of Node-Level Bias in GNNs.} Given a trained GNN $f_{\bm{\Theta}}$, a node $v_i$ to be explained, and its computation graph $\mathcal{G}_i = \{\mathcal{V}_i,  \mathcal{E}_i, \mathcal{X}_i\}$, our goal is to: (1) identify edges that are faithful to the prediction of $v_i$ (based on $f_{\bm{\Theta}}$) and maximally account for the bias exhibited in the GNN outcome of $v_i$; (2) identify edges that are faithful to the prediction of $v_i$ (based on $f_{\bm{\Theta}}$) and maximally contribute to the fairness level of the GNN outcome of $v_i$.
\end{myDef1}

Intuitively, Problem~\ref{p1} aims to identify two edge sets as two structural explanations: the bias explanation that accounts for the exhibited bias, and the fairness explanation that contributes to the fairness level of the given prediction. 
%
%
From the perspective of usability, the first explanation aims to identify edges that introduce the most bias to the instance-level GNN prediction, while the second explanation aims to identify edges that maximally contribute to the fairness level of the GNN prediction for any given node.


\begin{figure*}[]
    \centering
    \vspace{-5mm}
    \includegraphics[width=0.73\textwidth]{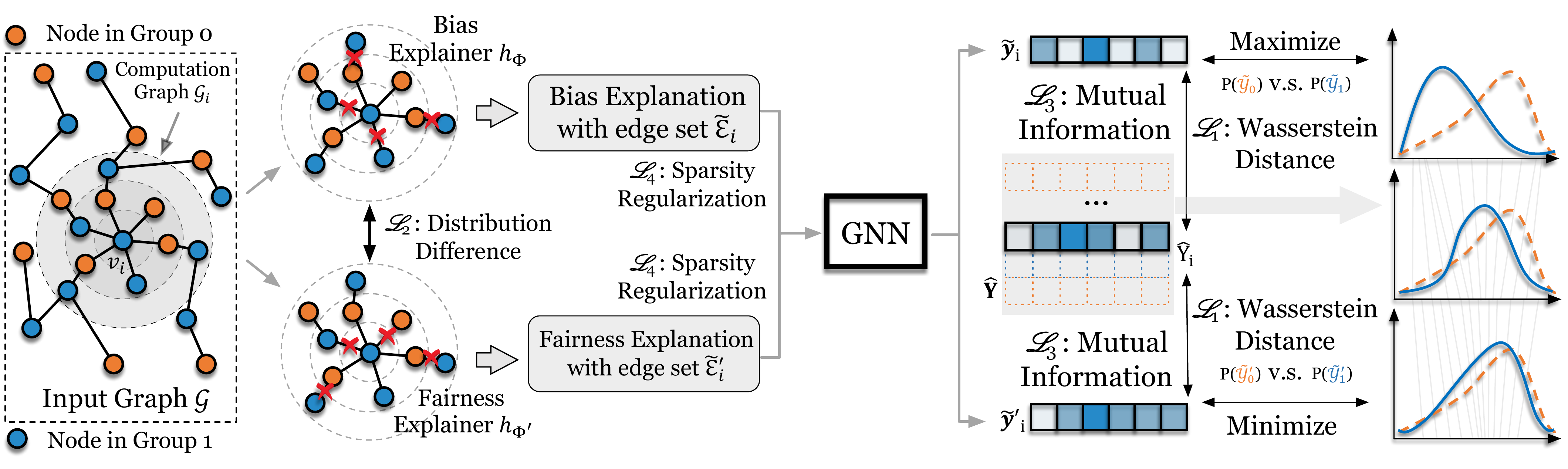}
    \vspace{-4mm}
    \caption{Framework structure of REFEREE: the edges in the edge set given by Bias Explainer maximally account for the node-level bias, while the edges in the edge set given by Fairness Explainer maximally alleviates the node-level bias.}
    \vspace{-3.5mm}
    \label{framework}
\end{figure*}

\section{The Proposed Framework}

\vspace{-0.3em}

In this section, we first present a principled metric to quantify the node-level bias for any given GNN prediction. Then we provide an overview of REFEREE, which is the proposed bias explanation framework for GNNs. Finally, we design a unified objective function for the proposed bias explanation framework REFEREE.


\subsection{Node-Level Bias Modeling}

\vspace{-0.3em}


To tackle the challenge of \emph{Fairness Notion Gap}, here we aim to formulate a novel metric to quantify the bias for the node-level GNN prediction.
Here we propose to formulate such a bias metric in the probabilistic outcome space of GNN predictions. The reason is that the information about the exhibited bias in the node-level prediction could be lost when the probabilistic outcomes are transformed into discrete predicted labels. In this regard, a bias metric based on the probabilistic outcome can better reflect the exhibited bias in the node-level predictions.
This is also in align with some existing bias measures~\cite{dai2021say,dong2021edits,fan2021fair}. However, although these existing bias metrics are defined in the probabilistic outcome space, they can only measure the level of bias for the predictions over the whole population, and thus cannot be directly grafted to our problem.

We introduce the rationale of our proposed bias metric for node-level GNN predictions as follows.
Intuitively, by measuring how much a node's outcome contributes to the overall bias on the whole population, we can have a better understanding of the bias level of this node's outcome. 
More specifically, assume the nodes can be divided into two sensitive subgroups based on the values of their sensitive features\footnote{Without loss of generality, we focus on binary sensitive attribute here.}.
The GNN outcome of nodes in the two sensitive subgroups forms two distributions, where the distance between the two distributions generally reflects the overall bias~\cite{fan2021fair,dong2021edits}.
For any specific node, if we change the probabilistic outcome of this node, the distribution distance between the outcome sets of the two sensitive subgroups will also change accordingly.
%
%
Ideally, if the outcome of a node has no contribution to the outcome distribution distance between the two sensitive subgroups, then the distribution distance cannot be further reduced no matter how the outcome of this node is changed.
In other words, a node that does not contribute to the overall bias should have an outcome with which the outcome distribution distance between the two sensitive subgroups is minimized.
Meanwhile, we can also employ the potential distribution distance reduction to measure the contribution of a node's outcome to the overall bias.
%
%
Based on such intuition, we then define \textit{Node-Level Bias in GNNs} as follows.

\begin{myDef}
\label{fair_def}
\textbf{Node-Level Bias in GNNs.}
Denote a probabilistic GNN outcome set as $\mathcal{\hat{Y}}$. 
Divide $\mathcal{\hat{Y}}$ into $\mathcal{\hat{Y}}_0$ and $\mathcal{\hat{Y}}_1$ as the outcome sets of the two demographic subgroups based on the sensitive feature.
Denote $D$ as the distance between the distributions of $\mathcal{\hat{Y}}_0$ and $\mathcal{\hat{Y}}_1$.
For node $v_i$, denote $D_{\text{min}}(i)$ as the minimum distance between the distributions of $\mathcal{\hat{Y}}_0$ and $\mathcal{\hat{Y}}_1$ by changing the value of $\mathbf{\hat{y}}_i \in \mathcal{\hat{Y}}$ while maintaining $\sum_{k=1}^{C} \mathbf{\hat{y}}_i[k] = 1$. We define $B_i = D - D_{\text{min}}(i)$ as the node-level bias of node $v_i$ for the GNN prediction.
%
\end{myDef}

Definition 1 introduces how to measure the bias exhibited in the node-level prediction given a trained GNN.
Clearly, the minimum value of $B_{i}$ is 0, i.e., if no change on the value of $\mathbf{\hat{y}}_i$ can be found to further reduce the distance between the distribution of $\mathcal{\hat{Y}}_0$ and $\mathcal{\hat{Y}}_1$, we say that node $v_i$ does not exhibit any node-level bias in the GNN outcome.
In this paper, we adopt the Wasserstein distance as the metric for distribution distance measurement, considering its superior sensitivity over other distance metrics~\cite{DBLP:journals/corr/ArjovskyCB17}. 
%
%
We will validate the consistency between Definition 1 and traditional fairness notions (e.g., \textit{Statistical Parity} and \textit{Equal Opportunity}) in Section~\ref{debiasing_exp}.

\subsection{Overview of Proposed Framework}
\vspace{-0.3em}

Here we present an overview of the proposed bias explanation framework for node-level GNN predictions.
In particular, to tackle the challenge of \textit{Usability Gap},  REFEREE is designed with two different explainers, i.e., a bias explainer $h_{\bm{\Phi}}$ and a fairness explainer $h_{\bm{\Phi}'}$. The two explainers aim to identify two different edge sets in the given computation graph as two structural explanations, i.e., the bias explanation and the fairness explanation. The two explanations are learned in a contrastive manner, in which way edges that account for different explanations can be better distinguished.
The basic goal of the bias explainer is to identify the edges that maximally account for the exhibited node-level bias, while the goal of the fairness explainer is to identify the edges whose existence can maximally alleviate the node-level bias for the instance-level GNN prediction.
Different GNN explanation models that are able to identify edge sets as the node-level explanations can be the backbone of the two explainers.
Besides, to reflect the true reasoning result in the given GNN model, both identified explanations should be faithful to the given GNN prediction. 
This leads to the challenge of \textit{Faithfulness Gap}: how to achieve the bias(fairness)-related goal of each explanation and maintain faithfulness to the given GNN prediction at the same time?
To tackle this challenge, we design a constraint to enforce the faithfulness of the identified explanations, and incorporate such constraint into a unified objective function for the proposed framework.
Optimizing such a unified objective function helps to achieve two goals: (1) the bias(fairness)-related explanation goals of both explainers; and (2) the goal of faithfulness through end-to-end training.
%
%
The overall structure of REFEREE is presented in Fig.~\ref{framework}.
%
Given a trained GNN $f_{\bm{\Theta}}$, a node $v_i$, and its computation graph $\mathcal{G}_i$, the goal of \emph{Bias Explainer} is to identify an edge set $\mathcal{\tilde{E}}_i$ that maximally accounts for the exhibited node-level bias of $v_i$ as the bias explanation, while the goal of \emph{Fairness Explainer} is to identify an edge set $\mathcal{\tilde{E}}_i'$ as the fairness explanation, where the edges in $\mathcal{\tilde{E}}_i'$ maximally alleviate the node-level bias of $v_i$.

\subsection{Objective Function}
\label{objectives}

\vspace{-0.3em}

In this subsection, we introduce the unified objective function formulation of our proposed framework REFEREE. Generally, the unified objective function includes three components, namely explaining bias (fairness), enforcing fidelity, and refining explanation.

\subsubsection{Explaining Bias (Fairness).}
%
%
Here we first introduce the bias (fairness)-related constraints to enable the two explainers to identify the edges that maximally account for the node-level bias and the edges whose existence maximally alleviate the node-level bias for a given GNN prediction, respectively.
We start from the constraint for the \emph{Bias Explainer}.
Given any computation graph, the basic goal of Bias Explainer is to identify the edges that maximally account for the node-level bias as an obtained edge set for the explanation.
Intuitively, if only edges that maximally account for the exhibited node-level bias are identified and preserved in such edge set, the probabilistic outcome based on such edge set will exhibit more node-level bias. 
This is because some edges whose existence help to alleviate the node-level bias in the vanilla computation graph are not involved in the obtained edge set anymore.
Based on such intuition, we develop the first component of our unified objective function towards the goal of explaining bias (fairness) as follows.

We denote the identified edge set given by Bias Explainer as $\mathcal{\tilde{E}}_i$. Here $\mathcal{\tilde{E}}_i \in \mathcal{\tilde{G}}_i$, and $\mathcal{\tilde{G}}_i$ is the computation graph with the obtained edge set $\mathcal{\tilde{E}}_i$. 
We represent the probabilistic outcome of the GNN model based on the computation graph $\mathcal{\tilde{G}}_i$ for node $v_i$ as $\tilde{\textbf{y}}_i = f_{\bm{\Theta}} (\mathcal{\tilde{G}}_i)$,
where $f_{\bm{\Theta}}$ is a given trained GNN model with fixed parameter $\bm{\Theta}$.
We utilize $\mathcal{\tilde{Y}}$ to denote the GNN outcome set $\mathcal{\hat{Y}}$ with the original element $\mathbf{\hat{y}}_i$ being replaced by $\mathbf{\tilde{y}}_i$, i.e., $\mathcal{\tilde{Y}} = \mathcal{\hat{Y}} \backslash \{\mathbf{\hat{y}}_i\} \cup \{\mathbf{\tilde{y}}_i\}$. 
According to the sensitive feature, we split $\mathcal{\tilde{Y}}$ into two outcome sets as $\mathcal{\tilde{Y}}_0$ and $\mathcal{\tilde{Y}}_1$ ($ \mathcal{\tilde{Y}}_0 \cup \mathcal{\tilde{Y}}_1 =  \mathcal{\tilde{Y}}$).
We denote the distribution of $\mathcal{\tilde{Y}}_0$ and $\mathcal{\tilde{Y}}_1$ as $P(\mathcal{\tilde{Y}}_0)$ and $P(\mathcal{\tilde{Y}}_1)$, respectively. 
Generally, if the vanilla probabilistic outcome $\mathbf{\hat{y}}_i$ is replaced with $\tilde{\textbf{y}}_i$, the outcome distribution distance between the two sensitive subgroups will also be changed accordingly. Considering that $\tilde{\textbf{y}}_i$ is derived based on the input computation graph $\mathcal{\tilde{G}}_i$, the identified edges in $\mathcal{\tilde{E}}_i \in \mathcal{\tilde{G}}_i$ then determine how the distribution distance changes.
As discussed above, the probabilistic outcome based on $\mathcal{\tilde{E}}_i$ will exhibit more node-level bias.
From Definition~\ref{fair_def}, we know that the identified edges in $\mathcal{\tilde{E}}_i$ are supposed to lead to a larger distribution distance between $P(\mathcal{\tilde{Y}}_0)$ and $P(\mathcal{\tilde{Y}}_1)$.
Correspondingly, we formulate the goal of bias explanation based on Wasserstein-1 distance as
%
\begin{align}
\label{ws_distance}  
    \max_{\mathcal{\tilde{E}}_i} \text{W}_1 (P(\mathcal{\tilde{Y}}_0), P(\mathcal{\tilde{Y}}_1)),
\end{align}
where $ \text{W}_1 (P(\mathcal{\tilde{Y}}_0), P(\mathcal{\tilde{Y}}_1))$ is formally presented as
\begin{align}
\begin{small}
\label{distance_expression}  
    W_1(P(\mathcal{\tilde{Y}}_0), P(\mathcal{\tilde{Y}}_1)) = 
    \inf \mathbb{E}_{ (\tilde{\mathbf{y}}_{(0)}, \tilde{\mathbf{y}}_{(1)}) \sim \kappa} [\| \tilde{\mathbf{y}}_{(0)} - \tilde{\mathbf{y}}_{(1)} \|_1].
\end{small}
\end{align}
\noindent Here $\kappa \in \Pi (P(\mathcal{\tilde{Y}}_0), P(\mathcal{\tilde{Y}}_1))$;
%
%
$\Pi(P(\mathcal{\tilde{Y}}_0), P(\mathcal{\tilde{Y}}_1))$ is the set including all possible joint distributions of $\kappa (\tilde{\mathbf{y}}_{(0)}, \tilde{\mathbf{y}}_{(1)})$ whose marginals are $P(\mathcal{\tilde{Y}}_0)$ and $P(\mathcal{\tilde{Y}}_1)$, respectively.
Generally, Eq.~(\ref{ws_distance}) encourages the Bias Explainer to identify edges that maximally account for the Wasserstein-1 distance between $P(\mathcal{\tilde{Y}}_0)$ and $P(\mathcal{\tilde{Y}}_1)$. 
Nevertheless, the infimum in Eq. (\ref{distance_expression}) is intractable. To perform effective optimization with gradient-based optimizing techniques (e.g., stochastic gradient descent), we adopted a widely used approximation strategy~\cite{cuturi2014fast} for the Wasserstein distance, which has been empirically proved to be effective~\cite{guo2020learning}.

We follow a similar approach to set up the other goal for Fairness Explainer to encourage the identification of edges whose existence can maximally alleviate the node-level bias for the given GNN prediction.
We assume the edge set given by Fairness Explainer as $\mathcal{\tilde{E}}_i'$, where $\mathcal{\tilde{E}}_i' \in \mathcal{\tilde{G}}_i'$. Here $\mathcal{\tilde{G}}_i'$ is the computation graph with the identified $\mathcal{\tilde{E}}_i'$.
We denote the outcome of the GNN model based on $\mathcal{\tilde{G}}_i'$ for node $v_i$ as $\tilde{\textbf{y}}_i' = f_{\bm{\Theta}} (\mathcal{\tilde{G}}_i')$.
We use $\mathcal{\tilde{Y}}_0'$ and $\mathcal{\tilde{Y}}_1'$ to denote the subsets of $\mathcal{\tilde{Y}}' = \mathcal{\hat{Y}} \backslash \{\mathbf{\hat{y}}_i\} \cup \{\mathbf{\tilde{y}}_i'\}$ according to the sensitive feature.
Correspondingly, $P(\mathcal{\tilde{Y}}_0')$ and $P(\mathcal{\tilde{Y}}_1')$ are the distributions of $\mathcal{\tilde{Y}}_0'$ and $\mathcal{\tilde{Y}}_1'$, respectively. 
We formulate the goal of Fairness Explainer as
\begin{align}
\label{ws_distance2}  
\min_{\mathcal{\tilde{E}}_i'} \text{W}_1 (P(\mathcal{\tilde{Y}}_0'), P(\mathcal{\tilde{Y}}_1')),
\end{align}
where $\mathcal{\tilde{E}}_i'$ is the edge set given by the Fairness Explainer for explanation.
To summarize, we formulate the objective function term towards explaining bias (fairness) as 
\begin{align}
\label{l3}  
    \mathscr{L}_1(\bm{\Phi}, \bm{\Phi}') = \text{W}_1 (P(\mathcal{\tilde{Y}}_0'), P(\mathcal{\tilde{Y}}_1')) - \text{W}_1 (P(\mathcal{\tilde{Y}}_0), P(\mathcal{\tilde{Y}}_1)).
\end{align}
The basic rationale is that when $\mathscr{L}_1$ is minimized, $\text{W}_1 (P(\mathcal{\tilde{Y}}_0), P(\mathcal{\tilde{Y}}_1))$ is maximized to encourage Bias Explainer to identify edges that account for the exhibited node-level bias; $\text{W}_1 (P(\mathcal{\tilde{Y}}_0'), P(\mathcal{\tilde{Y}}_1'))$ is minimized to encourage Fairness Explainer to identify edges whose existence can maximally alleviate the node-level bias.

Nevertheless, considering that the probabilistic outcome corresponding to only one explained node is changed during the optimization of Eq. (\ref{ws_distance}) (or Eq.~(\ref{ws_distance2})),
%
the numerical change of Wasserstein-1 distance could be small. Correspondingly, when using gradient-based techniques to optimize the two explainers in REFEREE, the gradients of $\mathscr{L}_1$ w.r.t. the learnable parameters in the two explainers could be similar. This could lead to a phenomenon that the two explainers tend to converge at similar solutions, which means that $\mathcal{\tilde{E}}_i \in \mathcal{\tilde{G}}_i$ and $\mathcal{\tilde{E}}_i' \in \mathcal{\tilde{G}}_i'$ could be close.
%
%
To better differentiate the edges that are supposed to be separated into two different explanations, here we propose to add a contrastive loss between the two explainers. The intuition here is to encourage the Bias Explainer and the Fairness Explainer to identify different edges from each other.
Specifically, the distribution difference between the edges in $\mathcal{\tilde{E}}_i$ and $\mathcal{\tilde{E}}_i'$ are maximized as an encouragement for identifying different edges. 
It should be noted that the edge sets given by both the two explainers (i.e., $\mathcal{\tilde{E}}_i$ and $\mathcal{\tilde{E}}_i'$) are based on the edge set $\mathcal{E}_i$ in the given computation graph. Correspondingly, we denote the distribution of $\mathcal{\tilde{E}}_i$ and $\mathcal{\tilde{E}}_i'$ conditional on the given $\mathcal{E}_i$ as $P_{\bm{\Phi}}(\mathcal{\tilde{E}}_i| \mathcal{E}_i)$ and $P_{\bm{\Phi}'}(\mathcal{\tilde{E}}_i' | \mathcal{E}_i)$, respectively.
We give the optimization problem as
\begin{align}
\label{KLD}  
    \max_{\bm{\Phi}, \bm{\Phi}'} \text{Dist\_Diff} (    P_{\bm{\Phi}'}(\mathcal{\tilde{E}}_i' | \mathcal{E}_i) \|  P_{\bm{\Phi}}(\mathcal{\tilde{E}}_i| \mathcal{E}_i) ),
\end{align}
%
%
Various metrics can be adopted as the distribution difference operator $\text{Dist\_Diff} (.)$, e.g., Jensen–Shannon divergence and Wasserstein distance, etc.
We give the second objective function term as
\begin{align}
\label{l4}  
    \mathscr{L}_2(\bm{\Phi}, \bm{\Phi}') = - \text{Dist\_Diff} (    P_{\bm{\Phi}'}(\mathcal{\tilde{E}}_i' | \mathcal{E}_i) \|  P_{\bm{\Phi}}(\mathcal{\tilde{E}}_i| \mathcal{E}_i) ).
\end{align}
Minimizing $\mathscr{L}_2$ helps to encourage the two explainers to yield different edge sets from each other as the identified explanations.

\subsubsection{Enforcing Fidelity.}
The explanations given by the two explainers should be able to reflect the true reasoning result given the node-level GNN prediction.
Hence, for both explainers (i.e., the Bias Explainer and Fairness Explainer), the output explanation should be faithful to the given GNN prediction. In other words, given a node $v_i$, the structural explanations given by both the two explainers should lead to the same predicted label based on the given GNN $f_{\bm{\Theta}}$.
%
%
%
%
%
Based on such intuition, here we leverage the mutual information between the original predicted label and the subgraph with the identified edge set for explanation to formulate fidelity enforcement. This also aligns with some existing works on GNN explanation~\cite{ying2019gnnexplainer}.
We first introduce the fidelity enforcement formulation for the Bias Explainer.
Specifically, for node $v_i$, the mutual information between the original GNN prediction $\hat{Y}_i \in \{1, ..., C\}$ and the underlying subgraph $\mathcal{\tilde{G}}_i$ is maximized to ensure that $\mathcal{\tilde{E}}_i \in \mathcal{\tilde{G}}_i$ encodes the critical information of the given GNN prediction, which is formulated as
\begin{align}
\label{mi_term}
    \max_{\mathcal{\tilde{G}}_i} \text{MI}(\hat{Y}_i, \mathcal{\tilde{G}}_i) = \text{H}(\hat{Y}_i) - \text{H}(\hat{Y}_i| \mathcal{\tilde{G}}_i).
\end{align}
Here MI($\cdot, \cdot$) denotes the mutual information computation operator, and H($\cdot$) represents the entropy function. 
%
It is worth mentioning that in Eq.~(\ref{mi_term}), the value of the entropy term $\text{H}(\hat{Y}_i) = \text{H}(f_{\bm{\Theta}} (\mathcal{G}_i))$ is fixed, as the explanation model is post-hoc (i.e., the parameters in the given GNN model are fixed).
Therefore, the optimization problem in Eq. (\ref{mi_term}) can be reduced to only minimizing the second entropy term, 
%
where $\text{H}(\hat{Y}_i| \mathcal{\tilde{G}}_i)$ can be presented as
\begin{align}
\label{entropy_term}  
    \text{H}(\hat{Y}_i| \mathcal{\tilde{G}}_i) = - \mathbb{E}_{\hat{Y}_i | \mathcal{\tilde{G}}_i} [\log P_{\bm{\Theta}}(\hat{Y}_i | \mathcal{\tilde{G}}_i)].
\end{align}
Considering fidelity is necessary for both explainers, we give the fidelity constraint for Fairness Explainer similarly. The objective function term to enforce fidelity for both explainers is given as
\begin{align}
\label{loss_1}  
\small
    \mathscr{L}_3 (\bm{\Phi}, \bm{\Phi}') = - \mathbb{E}_{\hat{Y}_i | \mathcal{\tilde{G}}_i} [\log P_{\bm{\Theta}}(\hat{Y}_i | \mathcal{\tilde{G}}_i)] - \mathbb{E}_{\hat{Y}_i | \mathcal{\tilde{G}}_i'} [\log P_{\bm{\Theta}}(\hat{Y}_i | \mathcal{\tilde{G}}_i')].
\end{align}
Minimizing $\mathscr{L}_3$ enforces the identified edges in $\mathcal{\tilde{G}}_i$ and $\mathcal{\tilde{G}}_i'$ to encode as much critical information to $\hat{Y}_i$ as possible.

\subsubsection{Refining Explanation.}
As mentioned in Section~\ref{intro}, our proposed explanation framework should be able to identify two edge sets, where the edges in one set can maximally account for the exhibited node-level bias, and the existence of the edges in the other set can maximally alleviate the node-level bias in GNNs.
%
%
Therefore, the identified explanations for both explainers should be maximally refined.
Intuitively, to refine the learned explanations, those goal-irrelevant edges for the GNN outcome of node $v_i$ should be maximally identified and removed from the structural explanations of both explainers, i.e., the learned edge sets from both explainers should be sparse.
Here we propose to regularize the sparsity of the identified edge set to remove those goal-irrelevant edges maximally.
We take the sparsity regularization of the Bias Explainer as an example.
Note that the explanation of the identified edge set is identified via a weighted mask matrix $\mathbf{M} \in \mathbb{R}^{|\mathcal{V}_i| \times |\mathcal{V}_i|}$ which indicates the edge importance score with entry values.
We propose to utilize the $\ell_1$-norm of the mask matrix $\mathbf{M}$ for the Bias Explainer as the regularization, i.e., $\| \mathbf{M}\|_1$. Considering both explainers, the corresponding objective function term $\mathscr{L}_4$ is formulated as 
\begin{align}
\label{loss_2}  
    \mathscr{L}_4(\bm{\Phi}, \bm{\Phi}') = \| \mathbf{M}\|_1 + \| \mathbf{M}'\|_1
\end{align}
for the two explainers. 
Here $\bm{\Phi}$ is the parameter of Bias Explainer $h_{\bm{\Phi}}$, and $\bm{\Phi}'$ denotes the parameter of Fairness Explainer $h_{\bm{\Phi}'}$.
$\mathbf{M}$ and $\mathbf{M}'$ are used to indicate the edge weights given by the explanations from the Bias Explainer and Fairness Explainer, respectively.
Besides, there are also cases where people are only interested in a certain number of top-ranked critical edges. In other words, there could be a pre-assigned budget $T$ for the explanation edge set $\mathcal{\tilde{E}}_i$, i.e., $|\mathcal{\tilde{E}}_i| \leq T$. 
%
In this case, we formulate the $\mathscr{L}_4$ as
\begin{align}
\label{loss_budget}  
    \mathscr{L}_4(\bm{\Phi}, \bm{\Phi}', T, T') = \text{ReLU}(\| \mathbf{M}\|_1 -T) + \text{ReLU}(\| \mathbf{M}'\|_1 -T')
\end{align}
given pre-assigned budget $T$ and $T'$ for $\mathcal{\tilde{E}}_i$ and $\mathcal{\tilde{E}}_i'$, respectively.
Intuitively, minimizing $\mathscr{L}_4$ helps to remove those goal-irrelevant edges maximally to refine the identified explanation.

\subsubsection{Unified Objective Function Formulation.}
Based on our discussions on enforcing fidelity, explaining bias (fairness), and refining explanation, we formally formulate the unified objective function for the proposed GNN explanation framework REFEREE as
\begin{align}
\label{l_total}  
    \mathscr{L} =  \mathscr{L}_1 +  \alpha \mathscr{L}_2 + \beta  \mathscr{L}_3 +  \gamma \mathscr{L}_4.
\end{align}
Here $\alpha$, $\beta$, and $\gamma$ are hyper-parameters controlling the effect of the three constraining terms. For any specific node to be explained, minimizing the objective function in Eq. (\ref{l_total}) aims to: (1) encourage the Bias Explainer to identify an edge set that maximally accounts for the node-level bias in the given GNN; and (2) encourage the Fairness Explainer to identify an edge set that maximally contributes to the fairness for the given GNN prediction.
%





\vspace{-0.2em}
\section{Experimental Evaluations}
\vspace{-0.3em}

In this section, we first introduce the downstream learning task and the real-world datasets adopted for evaluation. The experimental settings and the implementation details are then introduced. Next, we present the empirical evaluation results of our proposed framework from the perspective of \textit{Effectiveness of Explaining Bias (Fairness)}, \textit{Explanation Fidelity}, and \textit{Debiasing GNN with Explanations}.
In particular, we aim to answer the following research questions:
\textbf{RQ1:}  How well can REFEREE identify edges to explain bias (fairness) in GNNs given the prediction of a specific node?
\textbf{RQ2:}  How well can the explanations given by the two explainers in REFEREE be faithful to the given GNN?
\textbf{RQ3:}  How will the obtained explanations from REFEREE help with GNN debiasing for the whole population?

\vspace{-1.6ex}

\subsection{Experimental Settings}
\label{settings}

\vspace{-0.3em}


\subsubsection{Downstream Task \& Real-world Dataset.}
In this paper, we focus on the widely studied \textit{node classification} as the downstream task.
We adopt three real-world attributed networks for experiments -- German Credit, Recidivism, and Credit Defaulter~\cite{agarwal2021towards}, where all node labels are binary. A detailed description is in the Appendix.

\subsubsection{Explainer Backbones.}
Different GNN explanation approaches that are able to identify edge sets as the node-level explanations can be adopted as the backbone of the two explainers in REFEREE. To evaluate how well the proposed framework can be generalized to different explanation backbones, we adopt GNN Explainer~\cite{ying2019gnnexplainer} and PGExplainer~\cite{luo2020parameterized} as two backbones of explainers for evaluation.

\subsubsection{Baselines.}
To the best of our knowledge, no other work is able to give structural explanations for the exhibited node-level bias of GNNs. Therefore, we modify some existing GNN explanation approaches to adapt them to explain exhibited node-level bias in terms of the computation graph structure.
The adopted existing GNN explanation approaches for adaptation include the attention-based GNN explanation~\cite{velivckovic2017graph}, the gradient-based GNN explanation~\cite{velivckovic2017graph}, and two state-of-the-art GNN explanation approaches (GNN Explainer~\cite{ying2019gnnexplainer} and PGExplainer~\cite{luo2020parameterized}).
We elaborate more details on how we achieve the adaptation for these approaches as follows.
%

First, we introduce how we adapt these approaches as the baselines to evaluate \textit{Effectiveness of Explaining Bias (Fairness)}. For attention-based explanation, we directly add a bias(fairness)-related objective onto the vanilla loss function of a Graph Attention Network (GAT) model~\cite{velivckovic2017graph} to maximize (as Eq.~(\ref{ws_distance})) or minimize (as Eq.~(\ref{ws_distance2})) the Wasserstein-1 distance between the outcome distributions of the two sensitive subgroups. This enables the GAT model to identify the two types of critical edges for bias and fairness explanation, i.e., edges that maximally account for the exhibited node-level bias and edges whose existence can maximally alleviate the node-level bias. 
The learned attention weights are regarded as the indicator of the final explanations. For gradient-based explanation, we utilize the same objective function as the objective function adopted by attention-based explanation. The two types of critical edges for bias and fairness explanation are identified through gradient ascend w.r.t. the adjacency matrix of the given computation graph.
%
For GNN Explainer and PGExplainer, we modified their objective function in a similar way as the attention-based explanation. Specifically, a bias(fairness)-related objective is added onto the vanilla loss function for both explanation models. For any given computation graph, the two types of critical edges for bias and fairness explanation are identified through maximizing (as Eq.~(\ref{ws_distance})) or minimizing (as Eq.~(\ref{ws_distance2})) the Wasserstein-1 distance between the outcome distributions of the two sensitive subgroups.

Second, for the evaluation of \textit{Explanation Fidelity}, we aim to compare whether the GNN explanation backbones in REFEREE can still maintain their faithfulness to the given GNN prediction. Here the most widely-used GNN Explainer is adopted as the baseline model. Correspondingly, GNN Explainer is also adopted as the backbone of the two explainers in REFEREE for a fair comparison.

Third, for the evaluation of \textit{Debiasing GNNs with Explanation}, we adopt the same baselines as those adopted in the evaluation of \textit{Effectiveness of Explaining Bias (Fairness)}.

\subsubsection{Evaluation Metrics.}
We first introduce the metrics for the evaluation of \textit{Effectiveness of Explaining Bias (Fairness)}. 
Specifically, we evaluate how much the node-level bias $B_i$ is promoted or reduced between the two sensitive subgroups when only the identified edge set is utilized for the GNN prediction of the given node. Intuitively, this enables us to evaluate how well each explainer can identify those edges that maximally account for the exhibited bias and edges whose existence can maximally alleviate the node-level bias for the prediction, respectively.
For the evaluation of \textit{Explanation Fidelity}, a widely acknowledged metric is \textit{Fidelity}$-$ score~\cite{yuan2020explainability}. Traditionally, \textit{Fidelity}$-$ score measures the ratio of the consistent pairs between the vanilla correct predictions and the correct predictions based on the identified edge set. Nevertheless, to reflect the true reasoning process in GNNs, we argue that the faithfulness of those incorrect predictions is also critical, as bias may also exhibit and need to be explained for those incorrect predictions from the perspective of the usability of the GNNs. As a consequence, we extend the \textit{Fidelity}$-$ score to measure the ratio of the consistent pairs between all vanilla predictions and the predictions based on the identified edge set.
Formally, the extended fidelity metric for $M$ explained nodes can be measured with $\text { Fidelity }=\frac{1}{M} \sum_{i=1}^{M}\left(\mathds{1}\left(\hat{Y}_{i}=\tilde{Y}_{i}\right)\right)$.
%
Here $\hat{Y}_{i}$ represents the vanilla GNN prediction for node $v_i$.
$\tilde{Y}_{i}$ denotes the prediction of the given GNN $f_{\bm{\Theta}}$ for node $v_i$, where only the identified edges for explanation are preserved in the corresponding computation graph.
%
$\mathds{1}(\cdot)$ is the indicator function, which returns 1 if $\hat{Y}_{i}=\tilde{Y}_{i}$ and 0 otherwise.
%
Finally, for \textit{Debiasing GNNs with Explanation}, we utilize two traditional fairness metrics $\Delta_{SP}$ and $\Delta_{EO}$ to quantitatively evaluate how much the predictions of a GNN are debiased in terms of the whole population. Here $\Delta_{SP}$ and $\Delta_{EO}$ measure the positive prediction rate difference between two sensitive subgroups over all nodes and nodes with only positive class labels, respectively. Additionally, we use node classification accuracy to evaluate the GNN utility.



\vspace{-1.8ex}
\subsection{Effectiveness of Explaining Bias (Fairness)}
\vspace{-0.3em}

\begin{table*}[]
\vspace{-1mm}
\caption{$\Delta B_i$ (Promoted) and  $\Delta B_i$ (Reduced) present how much Wasserstein-1 distance between the outcome distribution of two sensitive subgroups improves and reduces on average.
Absolute values of normalized promotion and reduction are given in $\times 10^{-4}$ scale.
Larger values indicate better effectiveness in explaining bias (fairness). 
GE- and PGE- prefixes indicate the backbone of both explainers in REFEREE as GNN Explainer and PGExplainer, respectively.
The best results are in Bold.
}
\vspace{-3mm}
\label{edge_identification}
\centering
\small
\begin{tabular}{ccccccccc}
\hline
              & \multicolumn{2}{c}{\textbf{German}} &  & \multicolumn{2}{c}{\textbf{Recidivism}} &  & \multicolumn{2}{c}{\textbf{Credit}} \\
              \cline{2-3}  \cline{5-6}  \cline{8-9}
              & \textbf{$\Delta B_i$ (Promoted)}          & \textbf{$\Delta B_i$ (Reduced)}          &  & \textbf{$\Delta B_i$ (Promoted)}          & \textbf{$\Delta B_i$ (Reduced)}         &  & \textbf{$\Delta B_i$ (Promoted)}          & \textbf{$\Delta B_i$ (Reduced)}          \\
              \hline
\textbf{Att.}          &   6.11 $\pm$ 2.51           &  7.84 $\pm$ 3.48           &  &  4.58 $\pm$ 1.67           & 7.18 $\pm$ 2.24           &  & 6.72 $\pm$ 0.75             &  8.48 $\pm$  3.29           \\
\textbf{Grad.}         &   4.27 $\pm$ 0.98           &  5.60 $\pm$ 1.85           &  &  3.59 $\pm$ 2.02          & 4.42 $\pm$ 2.01           &  & 5.97 $\pm$ 1.07             & 9.79 $\pm$ 1.78            \\
\textbf{GNN Explainer} &   5.17 $\pm$ 1.20           &  3.37 $\pm$ 1.53           &  &  1.74 $\pm$ 0.72           & 3.55 $\pm$ 2.08           &  & 7.41 $\pm$ 1.75             & 9.24 $\pm$ 2.66            \\
\textbf{PGExplainer}   &   8.73 $\pm$ 0.74           &  9.37 $\pm$ 1.87          &  &  6.36 $\pm$ 2.39           & 8.66 $\pm$ 1.82           &  & 7.48 $\pm$ 2.70             & 10.54 $\pm$ 3.22            \\
\textbf{GE-REFEREE}     &   14.29 $\pm$ 2.73           &  \textbf{14.45 $\pm$ 2.29}           &  &  \textbf{13.94 $\pm$ 3.74}         & 12.05 $\pm$ 2.79           &  & 10.30 $\pm$ 2.64            & \textbf{15.07 $\pm$ 3.35}            \\
\textbf{PGE-REFEREE}    &   \textbf{15.72 $\pm$ 2.31}           &  11.97 $\pm$ 2.62           &  &  10.39 $\pm$ 4.08           & \textbf{12.57 $\pm$ 3.12}           &  & \textbf{11.57 $\pm$ 2.91}             & 14.67 $\pm$ 3.49            \\
\hline
\end{tabular}
\vspace{-1.4em}
\end{table*}

To answer \textbf{RQ1}, we compare our proposed framework REFEREE with other baselines to evaluate the effectiveness of explaining bias (fairness). Here we adopt the widely used model GAT as the trained GNN for experiments, and similar results can be observed based on other GNNs. Specifically, we first randomly sample 50 nodes to be explained (i.e., $M=50$). 
Then for each node, we obtain the two predictions of the given GNN $f_{\bm{\Theta}}$ based on the computation graph corresponding to each of the two identified edge sets given by the two explainers.
For obtained predictions, the normalized average value of how much the node-level bias $B_i$ is promoted (given by Bias Explainer) or reduced (given by Fairness Explainer) compared with the vanilla $B_i$ based on the complete computation graph is presented in Table~\ref{edge_identification}.
For both promotion and reduction of $B_i$, a larger value indicates better results, as more biased or fairer node-level outcome can be obtained based on the identified structural explanation.
We make the following observations from Table~\ref{edge_identification}:
\begin{itemize}[topsep=0pt]
    \item Stable promotion and reduction of node-level bias is observed in all GNN explanation approaches. This indicates that the Wasserstein distance-based objective functions formulated in Eq.~(\ref{ws_distance}) and Eq.~(\ref{ws_distance2}) effectively help to identify edges that account for the exhibited node-level bias and edges whose existence can alleviate the exhibited node-level bias.
    \item Existing GNN explanation models (e.g., GNN Explainer and PGExplainer) do not show any superior performance over other straightforward GNN explanation approaches such as Att and Grad. This observation implies that for these representative GNN explanation approaches, simply adding a constraint to explain bias (fairness) at the instance level only achieves limited effectiveness.
    %
    %
    \item Among all GNN explanation approaches, REFEREE yields the structural explanations that lead to the highest promotion and reduction of node-level bias in all datasets. Based on such observations, we argue that REFEREE achieves the best performance over other alternatives on identifying edges that account for the exhibited bias and whose existence can alleviate the exhibited node-level bias for the prediction. 
\end{itemize}

\vspace{-2ex}
\subsection{Explanation Fidelity}
\vspace{-0.3em}

\begin{table}[]
\caption{Explanation fidelity evaluation for different GNNs. 
%
%
Numerical results are in percentage. Vanilla denotes the explanation results given by the vanilla GNN Explainer. B. Explainer and F. Explainer represent the Bias Explainer and Fairness Explainer, respectively. The best results are in Bold.}
\vspace{-3mm}
\label{utility_table}
\centering
\small
\begin{tabular}{ccccc}
                     \hline
                     &              & \textbf{German} & \textbf{Recidivism} & \textbf{Credit} \\
                     \hline
\multirow{3}{*}{\textbf{GCN}} & \textbf{Vanilla}      & 88.02 $\pm$ 1.48      & 90.04 $\pm$ 1.43       & 85.26 $\pm$ 1.67     \\
                     & \textbf{B. Explainer}          & \textbf{92.20 $\pm$ 1.39}       & 90.26 $\pm$ 3.24       & 87.60 $\pm$ 2.79     \\
                     & \textbf{F. Explainer}          & 89.17 $\pm$ 0.85       & \textbf{92.08} $\pm$ 2.44       & \textbf{89.41} $\pm$ 4.08     \\
                     \hline
\multirow{3}{*}{\textbf{GAT}} & \textbf{Vanilla}      & 83.65 $\pm$ 3.02       & 87.91 $\pm$ 2.04       & \textbf{88.64 $\pm$ 3.41}     \\
                     & \textbf{B. Explainer}          & \textbf{85.71 $\pm$ 2.31}       & 90.51 $\pm$ 4.58       & 86.09 $\pm$ 2.07     \\
                     & \textbf{F. Explainer}          & 84.40 $\pm$ 1.57       & \textbf{91.98 $\pm$ 3.95}       & 87.04 $\pm$ 3.10     \\
                     \hline
\multirow{3}{*}{\textbf{GIN}} & \textbf{Vanilla}      & 88.58 $\pm$ 2.50       & \textbf{91.77 $\pm$ 1.42}       & 87.62 $\pm$ 2.60    \\
                     & \textbf{B. Explainer}          & 88.11 $\pm$ 1.78       & 90.26 $\pm$ 4.13       & 86.47 $\pm$ 2.13    \\
                     & \textbf{F. Explainer}         & \textbf{89.67  $\pm$ 2.23}       & 91.45 $\pm$ 1.78       & \textbf{88.17 $\pm$ 2.98}   \\
                     \hline
\end{tabular}
\vspace{-1.8em}
\end{table}

We then answer \textbf{RQ2} in this subsection. Generally, it is necessary to ensure that the structural explanation results given by both explainers in REFEREE are faithful to the given trained GNN, i.e., the identified edge sets should encode critical information for the given GNN predictions. More specifically, in our experiments, the predicted labels given by the GNN model based on the computation graph with the identified edge sets should be the same as those based on the vanilla computation graph.
%
%
To evaluate how well the proposed framework can maintain faithfulness when it is generalized to different GNNs, here we choose three widely used GNNs, namely GCN~\cite{DBLP:conf/iclr/KipfW17}, GAT~\cite{velivckovic2017graph}, and GIN~\cite{DBLP:conf/iclr/XuHLJ19} for explanation.
Fidelity is adopted as the metric for evaluation. Intuitively, fidelity measures to what proportion the predicted labels based on the identified explanation are maintained to be the same as the vanilla ones.
%
%
Here we adopt the GNN Explainer as the backbone of the two explainers in REFEREE.
For both the baseline model and our proposed framework, we train and make predictions five times separately for 50 randomly selected nodes. We present the performance comparison between the two explainers in our framework and vanilla GNN Explainer on the average performance of fidelity in Table~\ref{utility_table}.
%
We can make the observation that both Bias Explainer and Fairness Explainer achieve comparable performance on fidelity with the vanilla GNN Explainer across different datasets and GNNs. Consequently, we argue that the explanation given by REFEREE maintains faithfulness to the GNN predictions.


\begin{figure*}[!t]
	\centering 
	\vspace{-0.35cm}
    \subfloat[Changes of $\Delta_{SP}$]{
        \includegraphics[width=0.25\textwidth]{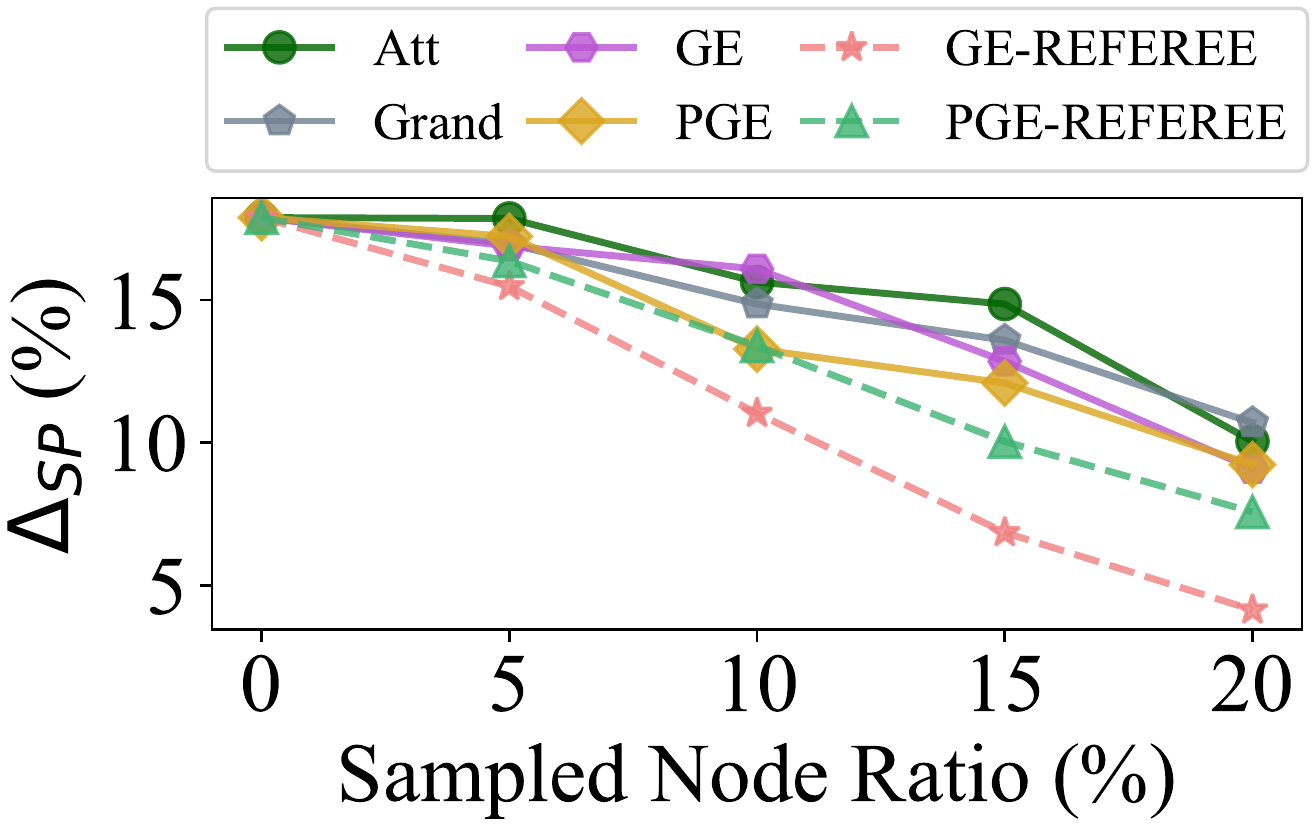}
    }
    \hspace{7mm}
        \subfloat[Changes of $\Delta_{EO}$]{
        \includegraphics[width=0.25\textwidth]{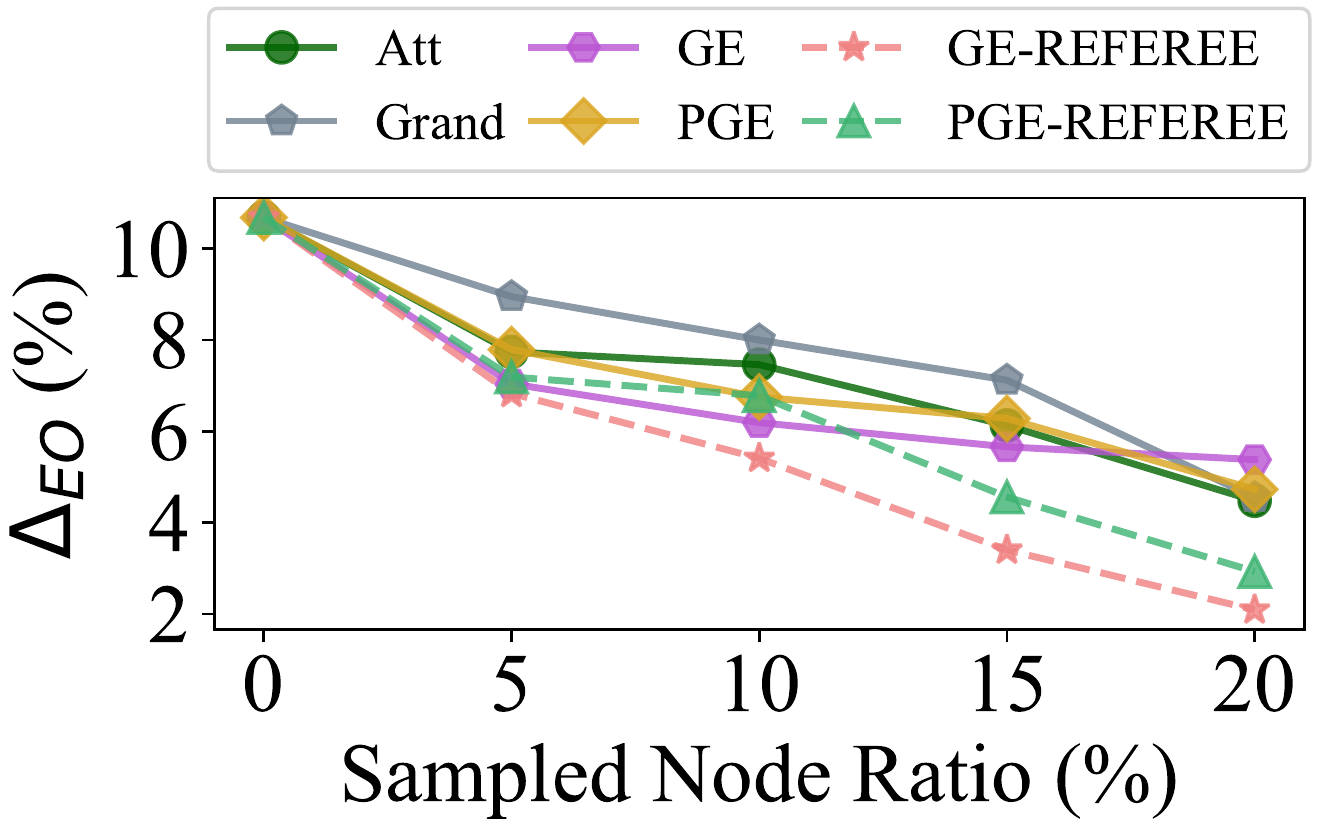}
    }
        \hspace{7mm}
    	\subfloat[Changes of accuracy]{
        \includegraphics[width=0.25\textwidth]{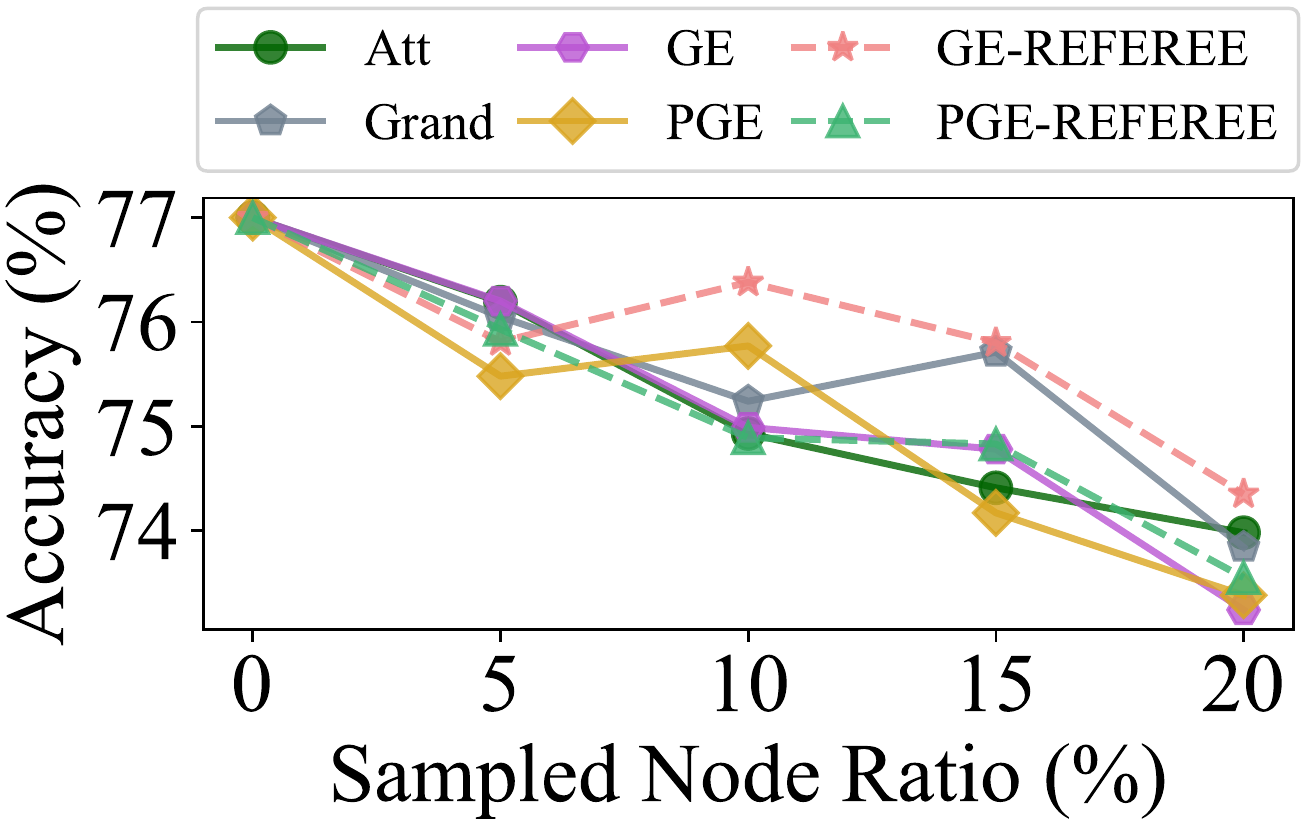}
    } 
    		\vspace{-4mm}
\caption{Debiasing GAT with explanations given by REFEREE with two different backbones and other baselines.}
	\vspace{-4mm}
	\label{tendency}
\end{figure*}

\vspace{-2ex}
\subsection{Debiasing GNNs with Explanations}
\label{debiasing_exp}

\vspace{-0.3em}

In this subsection, our goal is to answer \textbf{RQ3} to study how the instance-level explanations given by REFEREE help with GNN debiasing in terms of the whole population.
A straightforward approach here is to first identify the edges that tend to introduce bias in the outcome of GNNs for some randomly sampled nodes, then remove such bias-introducing edges and input the network data into the GNN model to obtain less biased predictions.
Nevertheless, the edges involved in the bias structural explanation (given by Bias Explainer) cannot be directly removed as a whole, as some edges could be critical for the GNN prediction of the explained node.
Besides, it is neither reasonable to only preserve the edges whose existence can maximally alleviate the node-level bias, as some removed non-critical edges for the explained node could be vital for the prediction of other nodes.
%
Here we adopt an alternative strategy to study how the explanations help with GNN debiasing in terms of the whole population.
Specifically, for those baseline explanation models, we randomly sample a subset of nodes for explanation. For each node, baselines are trained to learn structural explanations towards more biased and fairer predictions independently. Then edges that appear in the bias explanation but not in the fairness explanation are removed from the original input network.
The intuition here is that if an edge only appears in the edge set that maximally accounts for the exhibited bias but not in the edge set whose existence can maximally alleviate the node-level bias of the prediction, such edge can be regarded as being more critical to the exhibited bias instead of being more critical to an accurate and fair prediction. Therefore, removing edges bearing such property has the potential to reduce the exhibited bias while maintaining the utility (i.e., yielding accurate and fair predictions) of the GNN.
%
Correspondingly, for our proposed framework REFEREE, we also randomly sample nodes and remove edges that appear in the explanation given by Bias Explainer but not in the explanation from Fairness Explainer, i.e., removing edges in set $\mathcal{\tilde{E}}_i \backslash \mathcal{\tilde{E}}_i'$.
In this way, edges are removed from the input network data towards the goal of debiasing the GNN and maintaining its usability at the same time.
It is worth mentioning that such an edge removal strategy does not necessarily lead to graph structure modifications that are globally optimal for debiasing.
However, if fairer GNNs can be achieved via removing edges that exhibit node-level bias defined in Definition~\ref{fair_def}, the consistency between Definition~\ref{fair_def} and traditional fairness notions can be validated, i.e., reducing the node-level bias also helps to promote the overall fairness level of the GNN predictions in terms of traditional fairness metrics.

We adopt GAT as the explained GNN model here, and similar observations can also be found based on other GNNs. We vary the random sampling ratio of the number of explained nodes over the number of all nodes among \{0\%, 5\%, 10\%, 15\%, 20\%\}. The changes of node classification accuracy, $\Delta_{SP}$, and $\Delta_{EO}$ w.r.t. the sampled node ratio on German dataset is presented in Fig.~\ref{tendency}.
%
%
We make the following observations:
(1) With more nodes being sampled and more edges that only appear in the bias explanations being removed, both $\Delta_{SP}$ and $\Delta_{EO}$ reduce significantly. This verifies that removing the edges that account for the node-level bias generally alleviates the exhibited bias in terms of the whole population. Besides, the reduction of both $\Delta_{SP}$ and $\Delta_{EO}$ also validates the consistency between traditional fairness notions and node-level bias given in Definition~\ref{fair_def}. (2) Removing the edges that only appear in the bias explanations generally reduces the GAT prediction accuracy. We argue that it is because some edges that lead to more biased results could also be critical for accurate predictions. However, the accuracy reduction is within an acceptable range. (3) Compared with other baseline approaches, REFEREE leads to limited accuracy reduction but achieves a more significant reduction on $\Delta_{SP}$ and $\Delta_{EO}$. Such observation indicates that a fairer GNN is achieved (in terms of traditional fairness notions) based on the explanations identified by REFEREE compared with the explanations given by other alternatives. Considering our baselines also bear constraints for fidelity and explaining bias(fairness), it is safe to attribute such superiority to the designed contrastive mechanism of REFEREE. 
Consequently, we argue that REFEREE outperforms baselines in helping achieve fairer GNNs in terms of traditional fairness notions.

\vspace{-0.9em}
\section{Related Work}
\vspace{-0.3em}

\noindent \textbf{Explanation of GNNs.}
Generally, existing GNN explanation approaches can be divided into data-level approaches and model-level ones~\cite{yuan2020explainability}. 
For data-level approaches, the explanation models identify critical components in the input network data of GNNs, e.g., node features or edges.
For example, squared gradient values are regarded as the importance scores of different input features in the node classification task~\cite{baldassarre2019explainability};
interpretable surrogate models are leveraged to approximate the prediction of a certain GNN model, where the explanations from the surrogate model can be regarded as the explanation for the corresponding GNN prediction~\cite{huang2020graphlime,vu2020pgm}.
Another popular approach to identify important components of the input network data is to make perturbations on the input network, then observe the corresponding change in the output. The basic rationale is that if small perturbations lead to dramatic changes in the GNN prediction, then what has been perturbed is regarded as critical for the GNN prediction~\cite{ying2019gnnexplainer,luo2020parameterized,yuan2021explainability,schlichtkrull2020interpreting,wang2020causal}.
%
Despite its significance, this is a less studied topic. To provide model-level explanations for GNNs, graph generation can be leveraged to maximize the prediction of a GNN regarding a specific prediction (e.g., the probability of a class in graph classification)~\cite{yuan2020xgnn}. If the prediction probability of GNN regarding a specific prediction result can be maximized, then the generated input graph can be regarded as the explanation for this GNN that includes critical graph patterns.
Different from the existing GNN explanation approaches, our proposed framework REFEREE not only explores critical edges for GNN predictions, but also identifies their contribution to the bias in GNNs.
Hence, REFEREE is able to provide explanations for bias in GNNs, which helps understand how bias arises. This is with significance for GNN deployment in decision-critical scenarios and potentially facilitates the development of fairer GNNs.

\noindent \textbf{Fairness of GNNs.}
With the increasing societal concerns on the fairness of GNNs~\cite{wang2019semi}, explorations have been made to alleviate the bias exhibited in GNNs.
Generally, existing works focus either on \textit{group fairness}~\cite{dwork2012fairness} or \textit{individual fairness}~\cite{zemel2013learning}. Group fairness requires that GNNs should not yield biased predictions against any specific demographic subgroups~\cite{mehrabi2021survey}.
Among existing works, promoting group fairness through adversarial learning is one of the most popular GNN debiasing approaches~\cite{dai2021say}.
Its goal is to train a discriminator to identify the sensitive information from the learned node embeddings. When the discriminator can barely distinguish the sensitive feature given any learned embedding, the sensitive feature can be regarded as being decoupled from the learned embeddings.
Additionally, GNN debiasing can also be performed based on the input network data. For example, the network structure can be modified such that nodes in different demographic subgroups bear similar distributions on their neighbor node attribute values~\cite{dong2021edits}.
%
%
Moreover, edge dropout~\cite{spinelli2021biased} is also proved to be effective in debiasing GNNs.
On the other hand, individual fairness requires that similar individuals should be treated similarly~\cite{zemel2013learning,kang2020inform}.
However, promoting individual fairness for GNNs remains under-explored.
To the best of our knowledge, the only approach to fulfill such a goal is developed from a ranking perspective~\cite{dong2021individual}.





\vspace{-0.5em}
\section{Conclusion}
\vspace{-0.3em}

In this paper, we focus on a novel problem of structural explanation of node-level bias in GNNs. Specifically, we first propose to model node-level bias quantitatively, and then develop a principled post-hoc explanation framework named REFEREE with two different explainers: the bias explainer and the fairness explainer. 
Conditional on being faithful to the given GNN prediction, the two explainers aim to identify structural explanations that maximally account for the exhibited bias and that maximally contribute to the fairness level of the GNN prediction.
Experiments on real-world network datasets demonstrate the effectiveness of REFEREE in identifying edges that maximally account for the exhibited node-level bias and edges whose existence can maximally alleviate the node-level bias for any given GNN prediction.
Furthermore, REFEREE also shows superior performance over baselines on helping debias GNNs.



\vspace{-0.5em}
\section{ACKNOWLEDGMENTS}
\vspace{-0.3em}
This material is supported by the National Science Foundation (NSF) under grants No. 2006844 and the Cisco Faculty Research Award.

\vspace{-0.5em}

\bibliographystyle{ACM-Reference-Format}
\bibliography{reference}

\clearpage
\appendix
\section{Appendix}


\subsection{Reproducibility}

In this section, we focus on the reproducibility of our experiments as a supplement of Section~\ref{settings}. More specifically, we first present a detailed review of the three real-world datasets adopted in our experiments. Then we introduce the experimental settings, followed by the implementation details of our proposed framework REFEREE, GNNs, and baseline models. We finally present some key packages with the corresponding versions for our implementations.

\subsubsection{Real-World Datasets.}
Three real-world attributed network datasets are adopted in the experiments of this paper, namely German Credit, Recidivism, and Credit Defaulter~\cite{agarwal2021towards}. We present the statistics of these three datasets in Table~\ref{datasets}. Some detailed information corresponding to the three datasets is introduced as follows.
\begin{itemize} 
    \item \textbf{German Credit.} In German Credit dataset, nodes and edges represent the bank clients and the connections between client accounts, respectively. Here the gender of the bank clients is regarded as the sensitive feature, and the task is to classify if the credit risk of each client is high or not.
    
    \item \textbf{Recidivism.} In Recidivism, nodes represent defendants that were released on bail from the year 1990 to the year 2009, and edges represent the connection between defendants based on their past criminal records. Race of the defendants is considered as the sensitive feature, and the task is to classify if a certain defendant deserves bail. A positive bail decision indicates that the corresponding defendant is unlikely to commit a crime if released.
    
    \item \textbf{Credit Defaulter.} In Credit Defaulter, each node represents a credit card user, and an edge between two nodes is the connection between two credit card users. The age of the credit card users is the sensitive feature, and the task is to predict the future default of credit card payments for users.
\end{itemize}


\subsubsection{Experimental Settings.}
%
For the three real-world datasets used in this paper, we adopt the split rate for the training set and validation set as 0.8 and 0.1, respectively. The input node features are normalized before they are fed into the GNNs and the corresponding explanation models. For the downstream task \textit{node classification}, only the training set is available for all models during the training process. 
The trained GNN models with the best performance on the validation set are preserved for test and explanation.

\subsubsection{Implementation of REFEREE}
%
REFEREE is implemented in PyTorch~\cite{paszke2017automatic} and optimized through Adam optimizer~\cite{kingma2014adam}. In our experiments, the learning rate and training epoch number are set as 1e-5 and 200 for the explanation of all GNNs on all datasets.
Jensen–Shannon divergence is leveraged to measure the difference of the identified edge distributions between $\mathcal{\tilde{G}}_i'$ and $\mathcal{\tilde{G}}_i$, as its boundedness empirically leads to a more stable optimization process.
The edge budget $T$ (and $T'$) is set to be 500 and 200 for \textit{Effectiveness of Explaining Bias (Fairness)} and \textit{Debiasing GNNs with Explanations}, respectively.
In our experiments, hyper-parameter $\alpha$, $\beta$, and $\gamma$ are set as 1, 1e-4, and 1e-4, respectively. Numerical results for performance evaluation are based on the average of multiple runs.
Open-source code can be found at \href{https://github.com/yushundong/REFEREE}{https://github.com/yushundong/REFEREE}.

\subsubsection{Implementation of Graph Neural Networks.}
%
For all explained GNNs (i.e., GCN, GAT, and GIN) in our experiments, their released implementations are utilized for a fair comparison.
The layer number for GCN and GIN is set as 3. For GAT, we only adopt two layers due to the memory limit. The attention head number of GAT is set as one.
The hidden size is set as 20 for all explained GNNs.

\subsubsection{Implementation of Baselines.}
For all adopted GNN explanation baselines (i.e., Att, Grad, GNN Explainer, and PGExplainer), we also adopt their released implementations for a fair comparison. 
\begin{itemize}[topsep=0pt]
    \item \textbf{Att and Grad.} The implementations of Att and Grad are adopted based on the implementations of~\cite{ying2019gnnexplainer}.
    \item \textbf{GNN Explainer.} For GNN Explainer, we adopt the implementations of~\cite{ying2019gnnexplainer}. For the training of GNN Explainer, the learning rate is set as 0.001, and the weight decay rate is set as 0.005.
    \item \textbf{PGExplainer.} For PGExplainer, we adopt the implementations of~\cite{luo2020parameterized}. For the training of PGExplainer, the learning rate is set as 0.003.
\end{itemize}


\subsubsection{Packages Required for Implementations.}
%
We list the key packages and corresponding versions in our implementations as below. 
\begin{itemize}  
    \item Python == 3.7.10
    \item torch == 1.8.1
    \item torch-cluster == 1.5.9
    \item torch-geometric == 1.4.1
    \item torch-scatter == 2.0.6
    \item torch-sparse == 0.6.9
    \item cuda == 11.0
    \item numpy == 1.20.0
    \item tensorboard == 1.13.1
    \item networkx == 2.5.1
    \item scikit-learn == 0.24.1
    \item pandas==1.2.3
    \item scipy==1.4.1
\end{itemize}

\begin{figure}[htbp]
	\centering 
        \subfloat[The changes of relative $\Delta B_i$ (Reduced) w.r.t. $\alpha$ and $\beta$]{
        \includegraphics[width=0.48\textwidth]{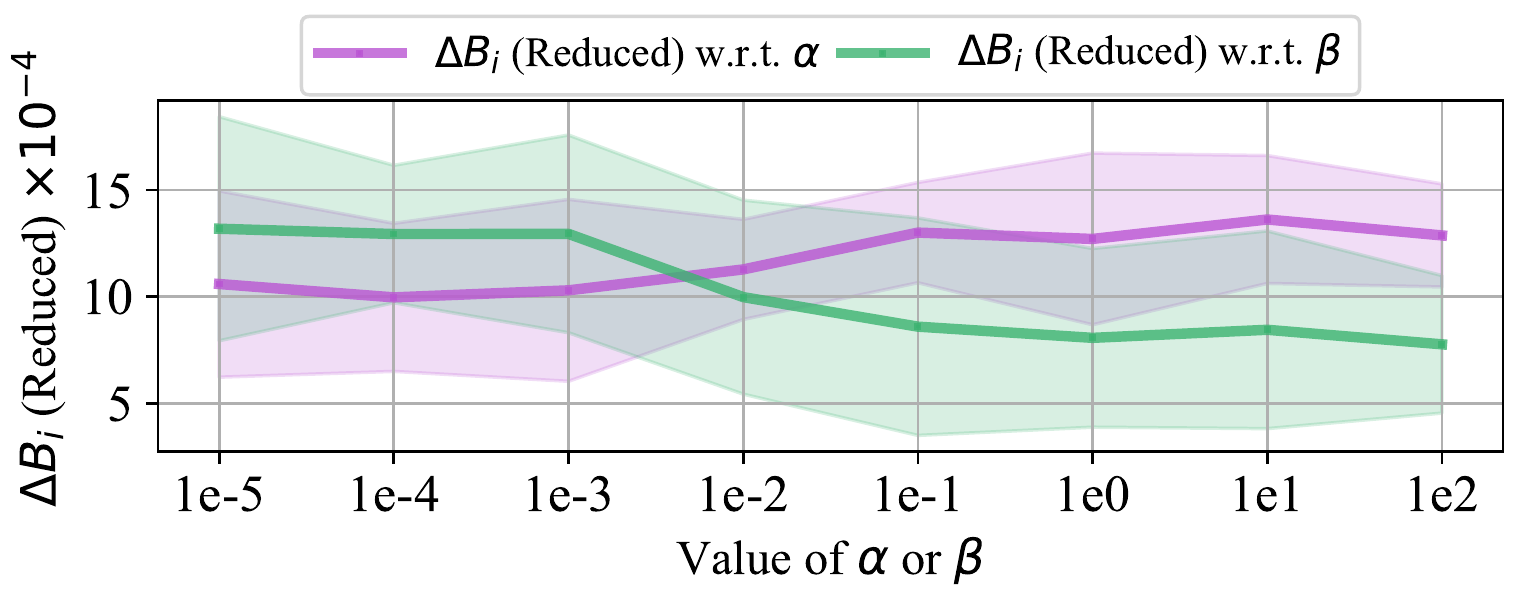}
        \label{param-1}
    } \\
    	\subfloat[The changes of $\Delta_{SP}$ w.r.t. $\alpha$ and $\beta$]{
        \includegraphics[width=0.48\textwidth]{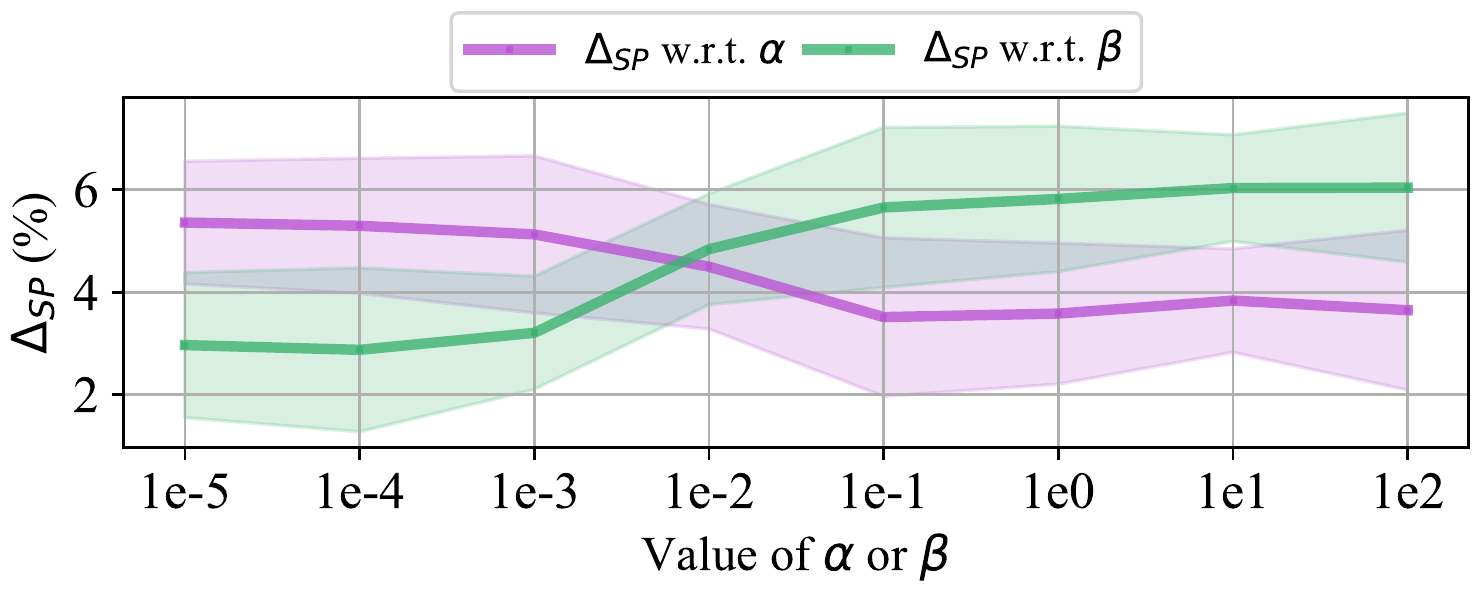}
                \label{param-2}
    }  \\
        \subfloat[The changes of accuracy w.r.t. $\alpha$ and $\beta$]{
        \includegraphics[width=0.48\textwidth]{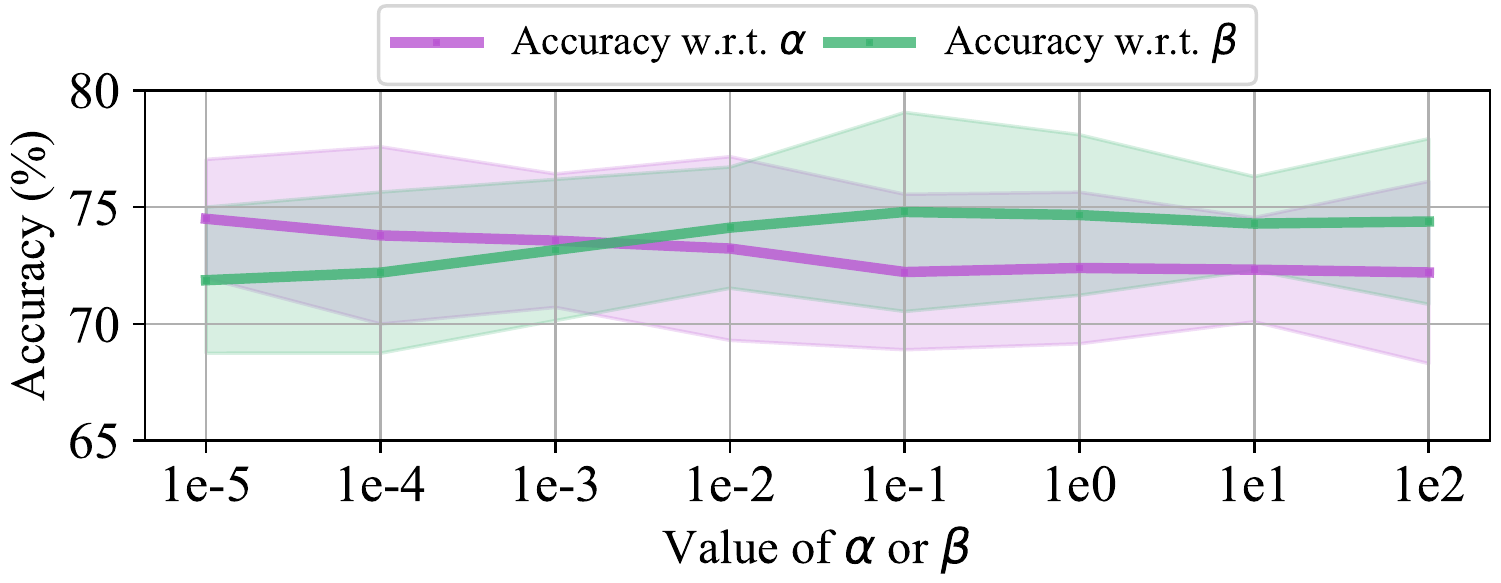}
                \label{param-3}
    }\\

    		\vspace{-3mm}
\caption{A parameter study of the proposed framework REFEREE based on hyper-parameter $\alpha$ and $\beta$. In (a), higher $\Delta B_i$ (Reduced) indicates better performance on explaining fairness for the Fairness Explainer in REFEREE. In (b), lower $\Delta_{SP}$ indicates a higher level of fairness is achieved based on the obtained explanations in terms of debiasing GNNs for the whole population. In (c), higher accuracy represents better GNN utility performance.}
	\vspace{-5mm}
	\label{param_study}
\end{figure}

\begin{table}[tbp]
\caption{Statistics of the three real-world graph data. \textit{Sens.} denotes the sensitive feature.}
\label{datasets}
\small
\centering
\begin{tabular}{lccc}
\hline
\hline
\textbf{Dataset}             & \textbf{German Credit}        & \textbf{Recidivism}   & \textbf{Credit Defaulter}         \\
\hline
\hline
\textbf{\# Nodes}               & 1,000                      & 18,876             & 30,000                         \\
\textbf{\# Edges}               & 22,242                     & 321,308            & 1,436,858                      \\
\textbf{\# Attributes} & 27                         & 18                 & 13                             \\
\textbf{Avg. degree}           & 44.5                      & 34.0              & 95.8                      \\
\textbf{Sens.} & Gender       & Race  & Age     \\
\textbf{Label}         & Good / Bad & Bail / No Bail   & Default / No Default  \\
\hline
\hline
\end{tabular}
\end{table}

\subsection{Summary of Notations}

To facilitate understanding, we present a summary of commonly utilized notations and the corresponding descriptions in Table~\ref{tb:symbols}.

\subsection{Parameter Sensitivity}
\label{sensitivity}

We present the parameter sensitivity of our proposed framework REFEREE in this section. More specifically, we explore how the hyper-parameters $\alpha$ and $\beta$ influence the performance of REFEREE on (1) explaining the bias (fairness) in GNNs and (2) debiasing the GNN across the whole population. 
Here $\alpha$ and $\beta$ control the effect of the distribution difference constraint between the two explanations from the two explainers and the constraint to achieve better fidelity, respectively.
In our experiments, we choose the widely used GAT model as the GNN to be explained, and we present the parameter study based on the performance of debiasing GNNs with explanations on German dataset. Similar observations can also be drawn on other GNN models and datasets.

\begin{table}[!t]
\small
\vspace{0.3cm}
\caption{Notations commonly used in this paper and the corresponding descriptions.} 
\vspace{-0.2cm}
\label{tb:symbols}
\begin{tabular}{cc}
\hline
\hline
\textbf{Notations}       & \textbf{Definitions or Descriptions} \\
\hline
\hline
$\mathcal{G}$   &  input graph\\
$\mathcal{V}$, $\mathcal{E}$, $\mathcal{X}$   &  node, edge set, and attribute set \\
$\mathcal{\hat{Y}}_i$           &  probabilistic GNN outcome set for all nodes \\
$\mathcal{G}_i$   &  computation graph centered on the $i$-th node\\
$\mathcal{\tilde{E}}_i$   &  bias explanation for the $i$-th node\\
$\mathcal{\tilde{E}}_i'$   &  fairness explanation for the $i$-th node\\
$\mathbf{\hat{y}}_i$           &  probabilistic GNN outcome of the $i$-th node \\
$\mathbf{\tilde{y}}_i$           &  outcome of the $i$-th node based on bias explanation\\
$\mathbf{\tilde{y}}_i'$           &  outcome of the $i$-th node based on fairness explanation\\
$\hat{Y}_i$           &  predicted label of the $i$-th node \\
$N$    & number of nodes in the input graph  \\
$M$    & number of nodes to be explained   \\
$C$    & number of classes for node classification   \\
\hline
\hline
\end{tabular}
\vspace{-0.4cm}
\end{table}

Now we introduce the experimental settings for the parameter sensitivity study. Specifically, we fix the value for parameter $\gamma$ as 1e-4 (the same as the setting in our implementation). 
First, for the parameter study of $\alpha$, we set $\beta=$1e-4 (the same as the setting in our implementation), and we vary $\alpha$ from \{1e-5, 1e-4, 1e-3, 1e-2, 1e-1, 1e0, 1e1, 1e2\}. 
Second, for the parameter study of $\beta$, we set $\alpha=1$ (the same as the setting in our implementation), and we also vary $\beta$ from \{1e-5, 1e-4, 1e-3, 1e-2, 1e-1, 1e0, 1e1, 1e2\}.
The performance changes of the proposed framework on explaining fairness (with $\Delta B_i$ (Reduced) being the node-level bias metric) and debiasing the GNN predictions over the whole population (with $\Delta_{SP}$ being the fairness metric and accuracy being the utility metric) are presented in Fig.~\ref{param_study}.
We can draw observations as below:
\begin{itemize}[topsep=0pt]
    \item From the perspective of explaining fairness (i.e., identifying the edges whose existence can maximally alleviate the exhibited node-level bias), we observe that a relatively larger $\alpha$ and a relatively smaller $\beta$ help to achieve better performance, i.e., larger $\Delta B_i$ (Reduced), in Fig.~\ref{param-1}. This is because larger $\alpha$ and smaller $\beta$ help the framework better differentiate the edges between the two types of explanations given by the two explainers. In this way, the fairness explainer is able to identify an edge set that leads to more significant node-level bias reduction, i.e., to give a fairness explanation that brings higher $\Delta B_i$ (Reduced) for any given node to be explained.

    \item From the perspective of debiasing the GNN predictions, we observe that a relatively larger $\alpha$ and a relatively smaller $\beta$ help to achieve better debiasing performance in Fig.~\ref{param-2}. This is because: (1) Larger $\alpha$ helps to better differentiate the edges between bias explanation and the fair explanation. This makes it easier for the framework to distinguish the edges that account for the exhibited bias and edges whose existence can alleviate the node-level bias. (2) Smaller $\beta$ means that the constraint strength on prediction fidelity is weak. This enables the framework to focus more on explaining bias (fairness) for edges in any given computation graph.
    
    \item From the perspective of maintaining GNN utility, we observe that a relatively smaller $\alpha$ and a relatively larger $\beta$ help achieve higher prediction accuracy in Fig.~\ref{param-3}. This is because smaller $\alpha$ and larger $\beta$ enforce the framework to focus more on the fidelity of the explanation. Therefore, more critical information could be encoded in the identified edges. Such an advantage leads to higher prediction accuracy based on the identified edges for any given node.
    
    \item Practically, it is necessary to balance the performance of bias reduction and model utility for any given GNN. In this regard, moderate values (e.g., values between 1e-4 and 1e0) for both $\alpha$ and $\beta$ are recommended.
\end{itemize}

\end{document}